\documentclass[runningheads]{llncs}

\usepackage{eccv}

\usepackage{eccvabbrv}
\usepackage[ruled,linesnumbered]{algorithm2e}

\usepackage{graphicx}
\usepackage{booktabs}

\usepackage[accsupp]{axessibility}  

\usepackage[pagebackref,breaklinks,colorlinks,citecolor=eccvblue]{hyperref}
\usepackage{multirow}
\usepackage{orcidlink}

\usepackage{xcolor}
\definecolor{mycolorpurple}{RGB}{255, 102, 255}
\definecolor{mycolorred}{RGB}{227, 26, 26}
\definecolor{mycolorskyblue}{RGB}{6, 185, 238}
\usepackage{threeparttable}

\begin{document}
	
	\title{M2DA: Multi-Modal Fusion Transformer Incorporating  Driver Attention for Autonomous Driving} 
	
	\titlerunning{Abbreviated paper title}
	
	\author{
		Dongyang Xu\inst{1,4}\orcidlink{0009-0008-9927-8649} \thanks{Corresponding author}
		\and
		Haokun Li\inst{2}\orcidlink{}
		\and
		Qingfan Wang\inst{1}\orcidlink{0000-0001-8556-1760}
		\and
		Ziying Song\inst{3}\orcidlink{0000-0001-5539-2599} 
		\and
		Lei Chen\inst{4}\orcidlink{} 
		\and
		Hanming Deng\inst{4}\orcidlink{0009-0005-6255-5479} 
		}

	\authorrunning{D.~Xu et al.}
	
	\institute{School of Vehicle and Mobility, Tsinghua University
		\and
		School of Mechanical Engineering, Beijing Institute of Technology
		\\
		\and
		School of Computer and Information Technology, Beijing Jiaotong University
		\\
		\and
		SenseTime Research
		\\
	\email{\{xudy22\}@mails.tsinghua.edu.cn}
}
	\maketitle

	\begin{abstract}	
		
		End-to-end autonomous driving has witnessed remarkable progress. However, the extensive deployment of autonomous vehicles has yet to be realized, primarily due to 1) inefficient multi-modal environment perception: how to integrate data from multi-modal sensors more efficiently; 2) non-human-like scene understanding: how to effectively locate and predict critical risky agents in traffic scenarios like an experienced driver. To overcome these challenges, in this paper, we propose a Multi-Modal fusion transformer incorporating Driver Attention (M2DA) for autonomous driving. To better fuse multi-modal data and achieve higher alignment between different modalities, a novel Lidar-Vision-Attention-based Fusion (LVAFusion) module is proposed. By incorporating driver’s attention, we empower the human-like scene understanding ability to autonomous vehicles to identify crucial areas within complex scenarios precisely and ensure safety. We conduct experiments on the CARLA simulator and achieve state-of-the-art performance with less data in closed-loop benchmarks. Source codes are available at  \href{https://anonymous.4open.science/r/M2DA-4772/}{M2DA}.
		
	\end{abstract}

	\section{Introduction}
	\label{sec:intro}
	
	With advancements in computational resources and artificial intelligence, significant progress has been made in autonomous driving. End-to-end autonomous driving methods map raw sensor inputs directly to planned trajectories, which are transformed into low-level control actions by applying control modules. Conceptually, this avoids the cascading errors inherent in complex modular designs and extensive manual rule-setting. Nonetheless, scalable and practical implementation of autonomous vehicles remains a substantial challenge. The primary obstacles include (1) how to enhance understanding of diverse environmental scenarios through multi-modal sensor integration and (2) how to improve scene understanding to effectively locate and predict critical risky agents in traffic scenarios.

	The camera-based TCP method \cite{wu2022trajectory} has shown remarkable performance on the Carla leaderboard, surpassing methods that rely on multi-modal sensor fusion, such as LAV \cite{chen2022learning} and Transfuser\cite{chitta2022transfuser}. This raises a question:\textit{ Does this imply that Lidar has become obsolete in end-to-end driving tasks?} Theoretically, the answer is no. Images provide detailed texture and color information but lack precise depth information. Conversely, point clouds offer accurate range views but have lower resolution. In theory, leveraging the complementary advantages of multi-modal data promises to enhance the performance of autonomous driving systems. According to the principle of information gain, adding more information should yield the same performance at the minimum, rather than leading to a performance degradation. So, \textit{what are the reasons behind the phenomenon mentioned above?}
	
	One underlying reason is the viewpoint misalignment caused by the ineffective fusion of point cloud and image information, eventually leading to the erroneous perception of the environment. For instance, misinterpreting or neglecting specific crucial data can lead to misjudgments of obstacles or inaccurate position estimation.
	
	Previous research about sensor fusion has predominantly focused on perception and prediction within the context of driving scenarios. This encompasses areas such as 2D and 3D object detection \cite{chen2017multi,chen2020mvlidarnet,zhou2020end,qi2018semi,ku2018joint,liang2019multi,meyer2019sensor} , along with motion prediction\cite{fadadu2022multi,liang2020pnpnet,luo2018fast,casas2020spagnn,casas2018intentnet,djuric2021multixnet,meyer2020laserflow}. These methods primarily leverage convolutional neural networks to learn and capture geometric and semantic information within 3D environments. However, such approaches either employ a locality assumption to align geometric features between the image and Lidar projection space or simply concatenate multi-sensor features. These fusion techniques may not effectively capture the interactions between multi-modal features in complex multi-agent scenarios.
	
	On the other hand, the highly dynamic, stochastic, and diverse characteristics of the traffic environment present a formidable challenge for autonomous driving. More specifically, autonomous vehicles should handle many unpredictable situations, such as vehicles disobeying traffic signals or pedestrians appearing suddenly from blind spots. Fortunately, in such intricate and hazardous environments, Proficient drivers are able to quickly identify and anticipate traffic dangers. For instance, they can subconsciously scan for oncoming traffic from all directions in unsigned intersections to avoid accidents preemptively.  Thus, driver attention (DA) can serve as a critical risk indicator, which has been proven to be effective in predicting driver behaviors or vehicle trajectories \cite{xing2019driver, martin2018dynamics, rangesh2020driver, gan2022multisource}. Meanwhile, experiments from naturalistic driving and in-lab simulator studies have consistently shown the effectiveness of DA in locating potential conflict objects, eventually enhancing road traffic safety \cite{dong2010driver, antin2019second, palazzi2018predicting, deng2019drivers}. 
	
	Consequently, accurately forecasting the focal points of a driver's gaze holds considerable significance for an end-to-end autonomous driving system to understand a complex traffic scenario. This predictive insight is instrumental in designing systems that can mirror human-like anticipatory skills, thereby enhancing the safety and reliability of autonomous vehicles. However, research on integrating DA into end-to-end autonomous driving has not been explored thus far.
	
	To overcome the above challenges, we propose a novel Multi-Modal fusion transformer incorporating Driver Attention (\textbf{M2DA}) framework for autonomous driving with two core innovations: efficient multi-modal environment perception and human-like scene understanding. First, to effectively capture the interactions between multi-modal features, we propose an innovative fusion module, Lidar-Vision-Attention-based Fusion (LVAFusion), that can integrate data from multi-modal and multi-view sensors. LVAFusion first utilizes global average pooling with positional encoding, which effectively encodes data from point clouds and images. By using these features as the query, LVAFusion can concentrate on the most relevant features within the context and highlight key features common to both sensor modalities, significantly improving the interpretation of their contextual interplay, compared to the methods that employ a randomly initialized query \cite{shao2023safety, hu2023planning, ye2023fusionad, prakash2021multi, chitta2022transfuser}. In addition, we also integrate driver attention into the framework to achieve human-like scene understanding. Upon forecasting the DA area within the current context, we treat it as a mask to adjust the weight of raw images to empower autonomous vehicles with the ability to effectively locate and predict risky agents in traffic scenarios like an experienced driver. Comprehensive experiments substantiate the effectiveness of incorporating the DA into end-to-end autonomous driving. To sum up, M2DA owns the following contributions:
	\begin{enumerate}
		\item  To avoid misalignment of critical objects across multiple modalities, we propose LVAFusion, a novel multi-modal fusion module, that utilizes queries with prior information to integrate image and point cloud representations. LVAFusion highlights key features common to both sensor modalities and captures their contextual interplay in a specific scenario.
		\item  To the best of our knowledge, we are the first to incorporate driver attention into end-to-end autonomous driving, which helps efficiently identify crucial areas within complex scenarios. The introduction of DA prediction not only provides finer-grained perception features for downstream decision-making tasks to ensure safety but also brings the scene understanding process closer to human cognition, thereby increasing interpretability.
		\item  We experimentally validate our approach in complex urban settings involving adversarial scenarios in CARLA. M2DA achieves state-of-the-art driving performance on both the Town05 Long benchmark and the Longest6 benchmark. 
	\end{enumerate}

	\section{Related Work}
	
	\subsection{End-to-end Autonomous Driving}
	Different from the traditional pipeline that is usually composed of different independent modules, such as object detection, motion prediction, and trajectory planning, the development of end-to-end autonomous driving systems without cumulative errors has become an active research topic in recent years, which has gained impressive driving performance, especially in closed-loop evaluation based on CARLA\cite{dosovitskiy2017carla}, a 3D driving simulation platform\cite{chen2020learning, chitta2021neat, codevilla2019exploring, hu2022model, ohn2020learning, sauer2018conditional, toromanoff2020end, zhang2021end, jia2023think, shao2023reasonnet,chen2021learning, zhang2023coaching,jaeger2023hidden,renz2022plant}. NEAT \cite{chitta2021neat} adopts neural attention fields to achieve efficient reasoning about the logical structure of traffic scenarios, especially in the dimensions of space and time. TCP \cite{wu2022trajectory} proposes an integrated approach combining trajectory planning and direct control methods in end-to-end autonomous driving, demonstrating superior performance in urban driving scenarios with a monocular camera input. Interfuser \cite{shao2023safety}, a safety-enhanced autonomous driving framework, addresses challenges related to comprehensive scene understanding and safety concerns by integrating multi-modal sensor signals and generating interpretable features for better constraint actions. To address imbalanced resource-task division, ThinkTwice \cite{jia2023think} adjusts the capacity allocation between the encoder and decoder and adopts two-step prediction (\ie, coarse-grained predicting and fine-grained refining) for future positions. Uniad \cite{hu2023planning} directly integrates full-stack driving tasks, including perception, prediction, and planning, into one unified network, effectively avoiding suffering from accumulative errors or deficient task coordination, which are common problems with traditional modular design methods. Note that most of the above models are trained via the Imitation Learning (IL) paradigm, which has intrinsic drawbacks; for example, their performance is limited by their rule-based teacher with privileged inputs. To improve this, Roach \cite{zhang2021end} adopts a Reinforcement Learning (RL)-based agent as the teacher, which demonstrates much better robustness. Its contemporaneous research, Latent DRL \cite{chen2021interpretable}, also trains an RL-based agent using a variational auto-encoder and generates intermediate features embedding from a top-down view image.
	
	Despite the impressive progress made by recent studies, we argue that there are still two aspects where current end-to-end autonomous driving can continue to improve: 1) more effective multi-modal environment perception that can better integrate data from multi-modal and multi-view sensors, 2) more human-like scene understanding that can quickly detect and predict critical risky agents in complex traffic scenarios like an experienced driver.
	
	\subsection{Sensor Fusion Methods for  Autonomous Driving}
	Owing to the complementary characteristic of different modalities, multi-modal sensor fusion has become a preferred approach across diverse research areas \cite{chen2017multi,chen2020mvlidarnet,zhou2020end,qi2018semi,ku2018joint,liang2019multi,meyer2019sensor, fadadu2022multi,liang2020pnpnet,luo2018fast,casas2020spagnn,casas2018intentnet,djuric2021multixnet,meyer2020laserflow}. For end-to-end autonomous driving, sensor fusion means the integration of heterogeneous data from diverse sensor types to refine the accuracy of perceptual information for autonomous driving, which provides an important foundation for subsequent safe and reliable decision-making. Recent methodologies in multi-modal end-to-end autonomous driving \cite{sobh2018end, xiao2020multimodal, behl2020label, natan2022end} reveal that the integration of RGB images with depth and semantic data can enhance driving performance. LAV \cite{chen2022learning} adopts PointPainting \cite{vora2020pointpainting} to fuse multi-modal sensor, which concatenates semantic class information extracted from the RGB image to the Lidar point cloud. ContFuse \cite{liang2018deep} exploits continuous convolutions to fuse image and Lidar feature maps at different levels of resolution. TransFuser \cite{chitta2022transfuser, prakash2021multi}, a widely used baseline model for CARLA, adopts multi-stage CNN to obtain multiple-resolution features and uses self-attention to process the image and Lidar representations independently, which fails to learn the complex correlation between different modalities. By contrast, cross-attention demonstrates more advantages in dealing with multi-modal features; thus, it is widely used in the recent SOTA works  (\eg, Uniad \cite{hu2023planning}, ReasonNet \cite{shao2023reasonnet} and Interfuser \cite{shao2023safety}). However, these approaches initialize the learnable queries of cross-attention with randomly generated parameters, which fails to utilize the prior knowledge buried in the multi-modal features. This might lead to the misalignment of the same critical object across multiple modalities, finally resulting in a slower and suboptimal convergence in model learning. To address this, we propose a novel multi-modal fusion method that uses cross-attention to interact image and Lidar representations, which is expected to achieve better alignment between different modalities.

	\subsection{Driver Attention Prediction}
	Human driver attention provides crucial visual cues for driving, so there has been a growing interest in predicting DA with various deep neural models recently \cite{tawari2017computational, huang2022temporally, lin2022ds, ma2022video, xie2022pyramid, tian2022driving, pang2020multi, qin2020u2, gan2022multisource, xia2020periphery, chen2023fblnet, deng2023driving, shi2023fixated}. \cite{lin2022ds, xie2022pyramid} use U-Net as the backbone and integrates the Swin-transformer to predict the DA. \cite{huang2022temporally} fuses transformer with a convolution network, then adopts a convlstm to process the features to predict DA. \cite{chen2023fblnet} proposes a feedback loop model, which attempts to model the driving experience accumulation procedure. \cite{gan2022multisource} proposes an adaptive model; it uses the domain adaption modules to predict DA in different traffic scenes. \cite{deng2023driving} utilizes a convlstm to capture the temporal features and employs a pyramid dilated convolution to extract spatial features. Then, they leverage an attention mechanism to fuse the temporal and spatial features and use these features to predict DA.  \cite{tian2022driving} proposes a dual-pathway model that enables a comprehensive analysis of both static and dynamic elements in driving environments. \cite{fu2023multimodal} proposes a multi-modal deep neural network that incorporates an anthropomorphic attention mechanism and prior knowledge for predicting DA. Despite so much research progress in DA prediction, there is still no research attempting to incorporate DA into end-to-end autonomous driving to gain excellent scene understanding ability from experienced human divers, which is addressed in this study.

	\section{M2DA }
\begin{figure}[tb]
		\centering
		\includegraphics[scale=0.0685]{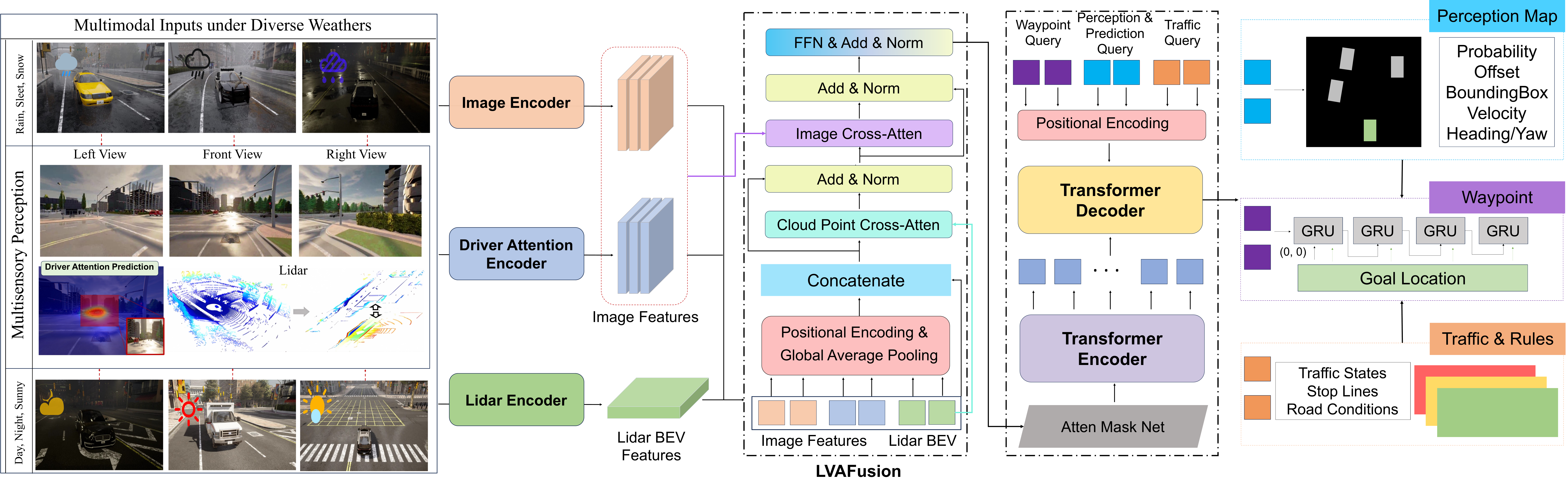}
		\caption{
			We present M2DA, a multi-modal fusion transformer incorporating driver attention, for end-to-end autonomous driving. M2DA takes multi-view images and Lidar cloud points as inputs. Firstly, we use a DA prediction model to mimic the focal points of drivers' visual gaze, which is treated as a mask to adjust the weight of raw images to enhance image data. Then, ResNet-based backbones are used to extract image features and Lidar BEV representations. We utilize global average pooling with positional encoding to encode these extracted representations. Then, they are treated as queries to calculate cross-attention with point clouds and images, respectively, and the outputs are considered as the final fused features, which are then fed into the subsequent transformer encoder. Three types of queries, \ie, waypoint query, perception and prediction query, and traffic query, are fed into the transformer decoder to obtain corresponding features for downstream tasks. Lastly, M2DA adopts an auto-regressive waypoint prediction network to predict future waypoints and uses MLPs to predict the perception map for surrounding objects and traffic states.
		}
		\label{fig:framework}
	\end{figure}

	An overview of M2DA is given in \cref{fig:framework}. M2DA consists of three main components: 1) a driver attention prediction module to mimic the focal points of the driver's gaze in current scenes; 2) LVAFusion using cross-attention to integrate data from multi-modal and multi-view sensors; 3) a transformer predicting the ego vehicle’s future waypoints and auxiliary information, such as the perception map for surrounding agents and traffic states. The following sections detail our problem setting, input and output representations, and each component of M2DA.
	
	\subsection{Problem Setting}
	
	\subsubsection{Task.} Given a start point and a target point, a series of sparse GPS coordinate target locations are calculated by a global planner. These waypoints guide M2DA navigation across diverse scenarios, such as highways and urban areas. Each route is initialized from predefined locations and contains multiple scenarios to test the agent's ability to handle different adversarial situations, such as making unprotected turns at intersections with a high density of various objects. The goal of the ego autonomous driving agent is to complete the route within a specified time while avoiding collisions and complying with traffic regulations.
	
	\subsubsection{Problem formulation.}  We adopt imitation learning to train our model, of which the goal is to learn a policy $\pi_\theta$ that imitates the expert behavior given the vehicle states at the current scene $\Pi$. It is comprised of multi-modal sensor input $I$, vehicle position in the global coordinate system $p$, vehicle speed $v$, and navigation information $n$. M2DA needs to output the future trajectory $\mathcal{W}$ and uses the control module to convert it into control signals $C$, including lateral control signals $steer \in \left[ -1, 1\right] $ and longitudinal control signals $brake \in \left[ 0, 1\right]$, $throttle \in \left[ 0, 1\right]$.
	
	We use a rule-based algorithm as our agent to collect the dataset, $\mathcal{D}=\left\{\left({\Pi}^i, \mathcal{W_{*}}^i\right)\right\}_{i=1}^Z$, where $Z$ is the size of the dataset. The collected expert trajectory $\mathcal{W_{*}}^i = \left\{\left(x_t, y_t\right)\right\}_{t=1}^T$ is defined in the 2D BEV space and is based on the ego-vehicle coordinate frame. The supervised training process of M2DA can be formulated as:
	\begin{equation}
		\arg \min _\theta \mathbb{E}_{\left(\mathbf{\Pi}, \mathcal{W}^*\right) \sim \mathcal{D}}\left[\mathcal{L}\left(\mathcal{W}_*, \pi_\theta({\Pi^*})\right)\right]
	\end{equation}
	where the multi-modal inputs $\Pi^*$ consist of Lidar point clouds, camera images from three perspectives, incorporating driver visual gaze information obtained from a pre-trained DA prediction model.  We use L1 distance to measure the loss between predicted trajectory $\pi_\theta({\Pi^*})$ and expert trajectory $\mathcal{W}_*$. Moreover, we add some auxiliary losses (perception and traffic states) to improve the performance, similar to \cite{shao2023safety, chitta2022transfuser}. Finally, we use two PID controllers to get the control signals $C = \Phi(\pi_\theta({\Pi^*})) $.
	
	\subsection{Input and Output Representations}
	\subsubsection{Input representations.} To better utilize the complementarity between cameras and Lidar, we use three RGB cameras ($60^o$ left, forward, and $60^o$ right) and one Lidar sensor. For Lidar point clouds, we follow previous works \cite{rhinehart2019precog, shao2023safety, chitta2022transfuser, gaidon2016virtual} to convert the Lidar point cloud data into a 2D BEV grid map by calculating the number of Lidar points inside each grid. The 2D BEV grid map area is set to 32×32m, with 28m in front of the vehicle, 4m behind the vehicle, and 16m to each side. We partition the grid into 0.125m × 0.125m cells, yielding a resolution of 256 × 256 pixels. For camera images, the setting is $100^o$ FOV and 800×600 resolution in pixels. Because of the distortion caused by the rendering of the cameras in the CARLA simulator, the images are cropped to 3 × 224 × 224. We use the DA prediction model to get the driver's gaze and consider it as a mask to modify the weight of raw images. Note that for all the sensor information, only data at the current time step is taken as inputs since previous researchers found that the integration of historical data does not invariably lead to an augmentation of performance for autonomous driving\cite{muller2005off, bansal2018chauffeurnet, wen2020fighting, wang2019monocular}. 
	
	\subsubsection{Output representations.} Inspired by \cite{shao2023safety}, M2DA produces two categories of outputs: safety-insensitive and safety-sensitive. For safety-insensitive outputs, M2DA predicts the future trajectory of the ego-vehicle in BEV space, represented by a sequence of 2D waypoints $\left\{\mathbf{w}_t=\left(x_t, y_t\right)\right\}_{t=1}^T$, where T is 10. The trajectory is then passed into two PID controllers to get the control signals. For safety-sensitive outputs, M2DA predicts the perception information of surrounding objects and traffic states as auxiliary tasks to avoid collisions or violations of traffic regulations. Concretely, the perception information of surrounding objects is represented by a heatmap image $M \in \mathbb{R}^{S \times S \times 7}$, where S is 20. It provides seven characteristics for potential objects in each grid (\ie, existence probability, offsets from ego vehicle on the x and y axes, width and length of the 2D bounding box, speed, and yaw).
	
	\subsection{Driver Attention Prediction} 
	DA prediction can provide the driver's visual gaze for the autonomous driving agent to enhance its ability to understand traffic scenarios like an experienced human driver. The DA prediction model in M2DA adopts an encoder-decoder architecture. For the encoder, we use MobileNet-V2 \cite{sandler2018mobilenetv2} as the backbone for quick prediction due to its small memory footprint and small FLOPs. A self-attention mechanism and a Gaussian kernel are used to process spatial features. Then, we adopt an inverted residual block to project these features and feed them into a gated recurrent neural network (GRU) with 128 hidden channels and kernel size 3 $\times$ 3 for sequence prediction. For the decoder, we utilize self-attention to process the features extracted by GRU. Three inverted residual blocks are employed to compress the channel dimensions for better feature representation. Additionally, we use another self-attention to enhance channel information. Finally, nearest-neighbor interpolation is adopted to upsample the features to the size of the input image, which is then smoothed by a convolution with a kernel size of 15×15.
	
	Since the agent will face various scenarios during driving, if the DA model does not have strong generalization ability, it may lead to wrong gaze points. To solve this problem, inspired by \cite{droste2020unified}, we use four datasets (details in supplementary, Appendix B) to train our DA prediction model and adopt a series of techniques, including domain-adaptive batch normalization (DABN), domain-adaptive priors, domain-adaptive smoothing, spatial attention, and channel attention. Following \cite{gan2022multisource}, DABN in our model can be represented as:
	\begin{equation}
		{DABN}^{t}(I_{f}^{t}, \alpha^{t}, \beta^{t}) = \alpha^{t} \left( \frac{I_{f}^{t} - \mu^{t}}{\sqrt{(\sigma^{t})^2 + \epsilon}} \right) + \beta^{t}
	\end{equation}
	\begin{equation}
		\mu^{t} = \frac{\sum_{c}^C \sum_{i}^H \sum_{j}^W I_{f}^{t}}{C \times H \times W} \quad \sigma^{t} = (\frac{\sum_{c}^C \sum_{i}^H \sum_{j}^ (I_{f}^{t} - \mu^{t})^2}{C \times H \times W})^{\frac{1}{2}}
	\end{equation}
	where the ${I}_{f}^t \in \mathbb{R}^{C \times H \times W}$ denotes front image features in town $t$, and $\alpha^{t}, \beta^{t}$ are learnable parameters. $\epsilon$ is a small constant to ensure  the stability of numerical calculation. 
		
	\subsection{LVAFusion: Attention based Fusion Module}
	\label{attention module}
	This study proposes a novel multi-modal fusion module, LVAFusion, to integrate data from multi-modal and multi-view sensors (\cref{fig:framework}).
	First, ResNet \cite{he2016deep} is used as the backbone of three perception encoders, \ie, image encoder, attention encoder, and Lidar encoder, to extract multi-view images features $ I_l, I_f, I_r$, driver attention features $ I_a$, and point cloud features $ I_{lidar}$, respectively. Then, these perception features are concatenated as a multi-modal feature ${I} \in \mathbb{R}^{\in d \times H \times W}$.
	
	To better capture the local semantic information embedded in one specific modality as well as the global semantic information coupled between multiple modalities, we define local sensor features and global sensor features for each modality. Regarding local features, LVAFusion processes data of a specific modality $\mathcal{I}_c$ with positional encoding (PE) and then adds it with a learnable view embedding $\zeta$. For global features, LVAFusion utilizes global average pooling to convert the $\mathcal{I}_c$  into $w \in \mathbb{R}^{d \times 1}$ and then adds it with a learnable sensor embedding $\vartheta$ and the view embedding $\zeta$. The above procedure can be formulated as:
	\begin{equation}
		\mathcal{K}_{\mathcal{I}_c} = Concat(\mathcal{K}_{local}, \mathcal{K}_{global})\quad  \quad
		\left\{
		\begin{aligned}
			\mathcal{K}_{local} &= \mathcal{I}_c + {PE}(\mathcal{I}_c) + \zeta \\
			\mathcal{K}_{global} &= w + \vartheta +\zeta
		\end{aligned}
		\right.
	\end{equation}
	where $\mathcal{I}_c  \in (I_l, I_{fa}, I_r, I_{lidar})$ means the feature of a specific modality, where $ I_{fa}$ represents the features that combining front-view image features $ I_{f}$ and driver attention features $ I_{a}$. Then, we concatenate these features from different modalities as:
	\begin{equation}
		\mathcal{K}_{concat} = {Concat}(\mathcal{K}_{{I}_l}, \mathcal{K}_{{I}_{fa}}, \mathcal{K}_{{I}_r},  \mathcal{K}_{{I}_{lidar}})
	\end{equation}
	
	We utilize two cross-attention mechanisms to process the concatenated features $\mathcal{K}_{concat} \in \mathbb{R}^{B_s \times N \times D_f}$, where $B_s$ denotes the batch size, $N$ is the number of tokens in the sequence, and $D_f$ represents the dimension of each token. First, the point cloud cross-attention layer takes point cloud features $\mathcal {K}_{I_{lidar}}$ as the key and value while taking $\mathcal{K}_{concat}$ as the query to obtain the intermediate features $\mathcal{K}_{inter}$, formulated as:
	\begin{equation}
		\mathcal{K}_{inter} = LN(\mathcal{K}_{concat} +  \text{softmax}\left(\frac{\mathcal{K}_{concat}\mathcal{K}_{I_{lidar}}^T}{\sqrt{D_l}}\mathcal{K}_{I_{lidar}}\right) )
	\end{equation}
	where $D_l$ denotes the dimension of the point clouds token. $LN$ represents the Layer Normalization. Second, the image cross-attention layer processes the intermediate features $\mathcal{K}_{inter}$ and $\mathcal{K}_{I_{image}}$ in a similar way:
	\begin{equation}
		\mathcal{K}_{fused} = LN(\mathcal{K}_{inter} +  \text{softmax}\left(\frac{\mathcal{K}_{inter}\mathcal{K}_{I_{image}}^T}{\sqrt{D_i}}\mathcal{K}_{I_{image}}\right) )
	\end{equation}
	where $\mathcal{K}_{I_{image}} = Concat(\mathcal{K}_{{I}_l}, \mathcal{K}_{{I}_{fa}}, \mathcal{K}_{{I}_r})$, and $D_i$ means the dimension of the image token. By using the multi-modal fusion features, \ie, $\mathcal{K}_{concat}$ and $\mathcal{K}_{inter}$ to query different single-modal features, \ie, $\mathcal {K}_{I_{lidar}}$ and $\mathcal{K}_{I_{image}}$, M2DA can concentrate on the most relevant features across different modalities, which is expected to improve the interpretation of their contextual interplay in complex traffic scenarios.

	\subsection{Transformer for Predicting Waypoints and Auxiliary Information} 
	As shown in \cref{fig:framework}, we pass the fused features $\mathcal{K}_{fused} \in\mathbb{R}^{B_s \times N \times D_f}$ into a transformer \cite{vaswani2017attention} to obtain the waypoints of the ego vehicle. Before this, we use some masks to process $\mathcal{K}_{fused}$ to enhance the generalization ability. The transformer encoder comprises $K$ stacked standard transformer encoder layers. In our work, $K$ is 6. Each layer consists of multi-head self-attention, two MLPs, and Layer Normalization \cite{ba2016layer}. The features processed by the transformer encoder are treated as the key and value in the cross-attention layer of the transformer decoder. In terms of cross-attention, three types of queries are designed: ${T}$ waypoint queries where $T$ is 10, $S^2$ perception and prediction queries where $S$ = 20, and one traffic state query.
		
	We feed the output of the transformer decoder $\mathcal{Z} \in \mathbb{R}^{B_s \times N \times D_f}$ into three parallel prediction modules that simultaneously forecast future waypoints of the ego, perception information of surrounding objects, and traffic states, respectively, enabling a concise yet comprehensive environmental interpretation for navigation. As for waypoint prediction, we pass $\mathcal{Z}_{wp} \in \mathbb{R}^{B_s \times N_{wp} \times D_f}$ into an auto-regressive network consisting of GRUs to predict the future sequence of 2D waypoints for the ego vehicle $\left\{\mathbf{w}_t=\left(x_t, y_t\right)\right\}_{t=1}^T$ following \cite{filos2020can, shao2023safety, chitta2022transfuser}. The GRU's direct outputs are regarded as increments, so we recover the exact positions by accumulation. In order to prevent the predicted waypoint sequence from deviating from the target location, we adopt a linear projection layer to embed the GPS coordinates of the target location into a 64-dimensional vector, which is taken as the initial hidden state of GRU. Regarding surrounding objects, an MLP block, consisting of two linear layers with a ReLU activation function, takes $\mathcal{Z}_{ht} \in \mathbb{R}^{B_s \times N_{ht} \times D_f}$ as inputs to predict the heatmap image $M \in \mathbb{R}^{S \times S \times 7}$ of surrounding objects. In terms of traffic states, $\mathcal{Z}_{tf} \in \mathbb{R}^{B_s \times N_{tf} \times D_f}$ is fed into a single linear layer to predict traffic lights and stop signs.
	
	\subsection{Controller}
	We use two PID controllers to obtain throttle, brake, and steer values from the predicted waypoint sequence $\left(x_t, y_t\right)_{t=1}^T$. For the PID controllers, we use the same configuration as \cite{shao2023safety}. Considering the complexity of the traffic system, autonomous vehicles can encounter safety-critical scenarios where a traffic light changes suddenly or a pedestrian crosses the road unexpectedly. In such scenarios, using waypoints alone to control autonomous vehicles may be unsafe. Thus, to further enhance safety, we adjust the vehicle control signals from the PID controllers by introducing a safety-heuristic method. The details can be see in supplementary (Appendix C).

	\subsection{Loss Functions} 
	
	For the stability of the training process, we first train the DA prediction model independently with the loss named $\mathcal{L}_{da}$. Then, we train the whole model of M2DA. Similar to \cite{shao2023safety, chitta2022transfuser, prakash2021multi}, we use $L1$ loss to calculate the errors between predicted waypoints and labels from the expert, named $\mathcal{L}_{wp}$. For auxiliary tasks, the perception errors of surrounding agents and traffic lights are represented as $\mathcal{L}_{ht}$ and $\mathcal{L}_{tf}$, respectively. More details are described in supplementary (Appendix A).
	
	\section{Experiments}
	
	M2DA is implemented on the open-source CARLA simulator with version 0.9.10.1. Please refer to supplementary (Appendix D) for implementation details. The following sections introduce the training dataset, benchmarks, evaluation metrics, the driving performance of M2DA, and ablation studies.

	\subsection{Data Collection}
	M2DA is based on an imitation learning framework that requires an expert to collect driving data. Thus, we use a rule-based driving algorithm as the expert with 3 cameras (front, left, right), a Lidar, an IMU, a GPS, and a speedometer. We run the rule-based expert agent on all eight towns with different weather and collect 200K frames of driving data with 2$H_{z}$ due to the limitation of hard-disk capacity.

	\subsection{Benchmark and Metrics}
	We conduct experiments on two widely used benchmarks in CARLA, \ie, {Town05 Long} and {Longest6}, to conduct closed-loop autonomous driving evaluations. Details of the benchmarks can be seen in supplementary (Appendix E). In these benchmarks, the ego vehicle is required to navigate along predefined routes while ensuring no collisions occur and adherence to prevailing traffic regulations in the presence of adversarial conditions.
	
	We use three metrics introduced by the CARLA LeaderBoard to evaluate our framework: {Route Completion (RC)} is the percentage of the route distance completed. {Infraction Score (IS)} is a penalty coefficient representing the number of infractions made along the route. {Driving Score (DS)}, the most critical comprehensive metric, is the product of Route Completion and Infraction Score.

	\subsection{Comparsion with SOTA}
	We compare M2DA with state-of-the-art approaches in Town05 Long benchmark and Longest6 benchmark. Due to the randomness of the CARLA traffic manager and sensor noises, the evaluation results demonstrate a level of uncertainty. Thus, we repeat each evaluation experiment three times and report the average results. 
	
	For the Town05 Long benchmark (\cref{town5result}), our method achieves the best performance with DS of 72.6 and IS of 0.80, meaning that M2DA can handle complex scenarios well and reduce the occurrence of infractions. Some SOTA methods, \eg, MILE and DriveAdapter, obtain higher RC; however, they exhibit a significantly higher incidence of collisions or traffic violations. For Transfuser and Interfuser that use the same sensor configuration as M2DA, our model outperforms Transfuser in all metrics and only performs worse than Interfuser in RC. The results of Longest6 benchmark are shown in supplementary (Appendix F). 
	
	\cite{jaeger2023hidden} proved that the size of the collected expert data has a marked impact on driving performance. Despite being trained on a dataset of only 200K frames, M2DA outperforms existing state-of-the-art models using significantly larger training datasets, such as Interfuser (3M), MILE (2.9M), and Thinktwice (2M) on the Tonw05 benchmark, implying that M2DA can attain superior performance with a reduced amount of data.
	
	\begin{table}[tbp]
		\centering
		\caption{Comparison of M2DA with several state-of-the-art methods in the Town05 Long benchmark. $\uparrow$ means the higher, the better. C represents camera, and L means Lidar. Extra supervision refers to additional labels needed for training, apart from the actions and states of the ego vehicle. Expert denotes the extraction of knowledge from privileged agents. Box refers to the bounding box of other agents. The evaluation of DriveAdapter only runs once, denoted by superscript 1.}
		\resizebox{\textwidth}{!}{
			\begin{tabular}{lllllccc}
				\toprule
				Method & Fusion  & Modality & Extra Supervision & Dataset & DS $\uparrow$ & RC $\uparrow$ & IS $\uparrow$ \\
				\midrule
				CILRS \cite{codevilla2019exploring} & ResNet + Flatten  & C1 & None & \quad- & $7.8 \pm 0.3$ & $10.3 \pm 0.0$ & $0.75 \pm 0.05$ \\
				LBC \cite{chen2020learning} & ResNet + Flatten  & C3 & Expert & 157K & $12.3 \pm 2.0$ & $31.9 \pm 2.2$ & $0.66 \pm 0.02$ \\
				Transfuser \cite{chitta2022transfuser} & Fusion via Transformer  & C3L1 & Dep+Seg+Map+Box & 228K & $31.0 \pm 3.6$ & $47.5 \pm 5.3$ & $0.77 \pm 0.04$ \\
				Roach \cite{zhang2021end} & ResNet + Flatten  & C1 & Expert & \quad- & $41.6 \pm 1.8$ & $96.4 \pm 2.1$ & $0.43 \pm 0.03$ \\
				LAV \cite{chen2022learning} & PointPaiting & C4L1 & Expert+Seg+Map+Box & 189K & $46.5 \pm 2.3$ & $69.8 \pm 2.3$ & $0.73 \pm 0.02$ \\
				TCP \cite{wu2022trajectory} & ResNet + Flatten  & C1 & Expert & 189K & $57.2 \pm 1.5$ & $80.4 \pm 1.5$ & $0.73 \pm 0.02$ \\
				MILE \cite{hu2022model} & ResNet + Flatten  & C1 & Map+Box & 2.9M & $61.1 \pm 3.2$ & $\textbf{97.4 }\pm 0.8$ & $0.63 \pm 0.03$ \\
				Interfuser \cite{shao2023safety} & Fusion via Transformer  & C3L1 & Box & 3M & $68.3 \pm 1.9$ & $95.0 \pm 2.9$ & -- \\
				ThinkTwice \cite{jia2023think} & Geometric Fusion in BEV  & C4L1 & Expert+Dep+Seg+Map & 2M & $70.9 \pm 3.4$ & $95.5 \pm 2.6$ & $0.75 \pm 0.05$ \\
				DriveAdapter$^{1}$ \cite{jia2023driveadapter} & Geometric Fusion in BEV  & C4L1 & Expert+Seg+Map & 2M & $71.9 $ & $97.3 $ & $0.74 $ \\
				\midrule
				\textbf{M2DA} (ours) & LVAFusion  & C3L1 & Box & 200K & $\textbf{72.6} \pm 5.7$ & $89.7 \pm 7.8$ & $\textbf{0.80} \pm 0.05$ \\
				\bottomrule
			\end{tabular}
		}
		\label{town5result}
	\end{table}

	\subsection{Visualizations}
\begin{figure}[tb]
		\label{pic1}
		\centering
		\includegraphics[scale=0.32]{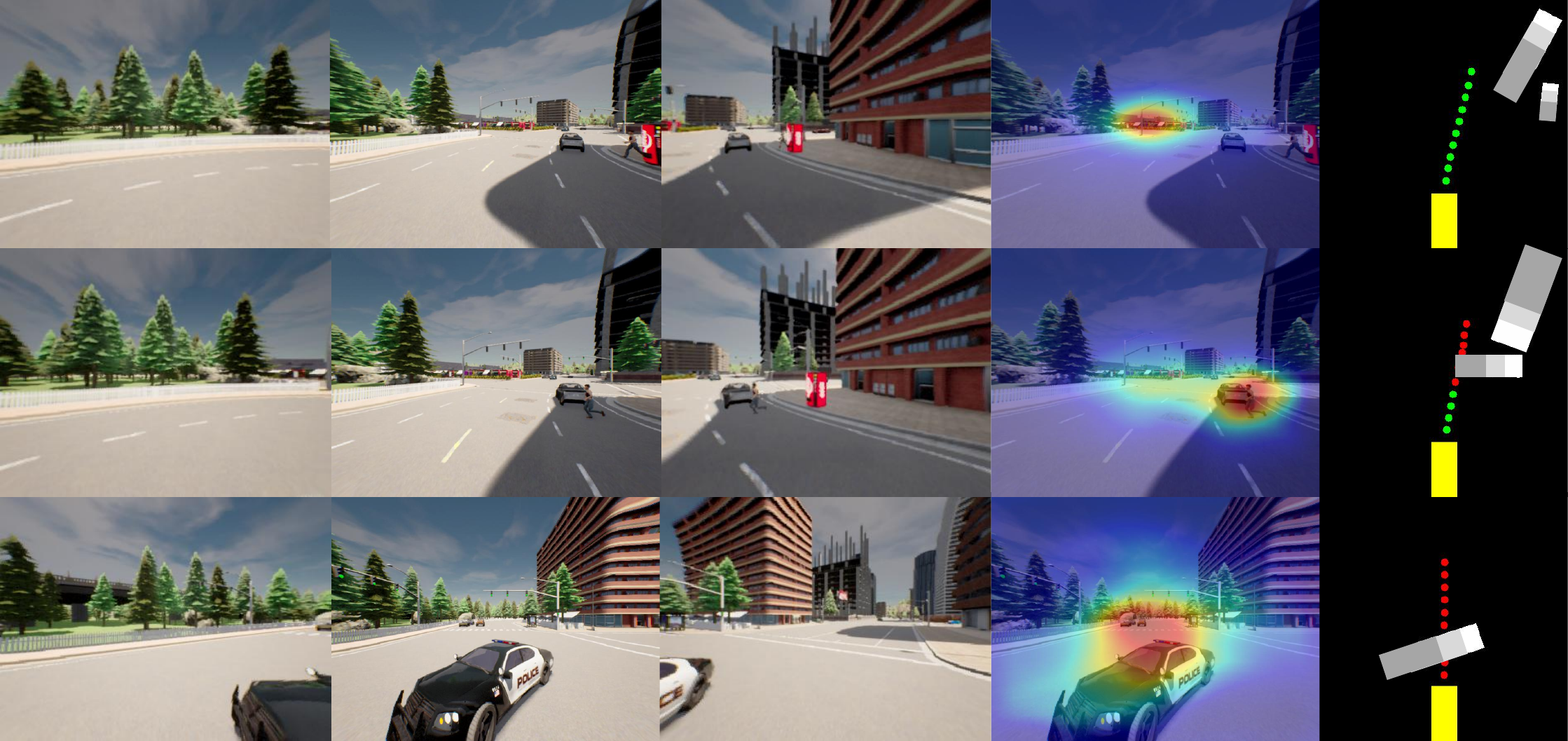}
		\caption{Each row represents a representative traffic scenario encountered by M2DA. The three columns on the left display the left-view, front-view, and right-view images, respectively. The fourth column shows the prediction results for driver attention. The last column represents the perceived states of surrounding vehicles. The yellow box denotes the ego vehicle. White, light gray, and gray boxes represent the perceived surrounding vehicles' current positions, predicted positions at the next time interval, and predicted positions at the next two time intervals, respectively. Green dots and red dots represent safe future trajectories of the ego and unsafe areas where collisions are likely to occur, respectively.
		}
		\label{fig:pic1}
	\end{figure}

	We visualize some representative cases in the evaluation results of M2DA (\cref{fig:pic1}). The first row displays a normal traffic scenario without apparent risks, where M2DA located its visual attention at the road vanishing point in the center of the image. In the second row, a running pedestrian was about to cross the road. In such a sudden situation, like an experienced human driver, M2DA quickly and accurately captured the dangerous object, \ie, the pedestrian, in the current traffic scenario and made corresponding driving decisions to avoid potential collisions. In a more dangerous scenario depicted in the third row, M2DA also quickly allocates attention to the vehicle at an intersection. Meanwhile, considering the predicted future trajectories of the vehicle, M2DA perceived a high risk of collision and instantly initiated emergency braking maneuvers to prevent accidents. Other visualizations can be seen in supplementary (Appendix F).
	
	\subsection{Ablation Studies}
	
	We now analyze several design choices for M2DA in a series of ablation studies on the Town05 Long benchmark.
	
	We first investigate the effect of different sensor modalities by utilizing different combinations of sensor inputs. Results are shown in \cref{ablationsensor}. 1C utilizes only the front RGB images as inputs, making it challenging to detect obstacles on the sides of the ego vehicle. Consequently, it exhibits the highest collision rate with vehicles (Veh) and the longest timeout (TO), eventually leading to the worst driving performance. When the left and right cameras are incorporated, 3C can observe traffic conditions more comprehensively, which not only reduces the risk of vehicle collisions but also alleviates the timeout. However, only taking camera images as inputs, 3C still demonstrates a high rate of running red lights (Red), indicating that the agent struggles to capture traffic light information effectively. To address this issue, we introduce driver attention as an additional input and make the model learn from experienced human drivers to allocate visual attention at traffic lights in advance when approaching a signalized intersection. As a result, 3C1A exhibits a lower rate of running red lights, leading to an increase in IS and DS. The further introduction of Lidar point clouds further improves IS, resulting in the highest DS.
	
	\begin{table}[tb]
		\centering
		\caption{Ablation study for different sensor inputs. $\uparrow$ means the higher, the better, while $\downarrow$ represents the lower, the better. 1C and 3C represent using one camera (front) and three cameras (left, front, right) as inputs, respectively. 3C1A means three cameras combining driver attention features. 3C1A1L further introduces one Lidar.}
		\resizebox{\textwidth}{!}{
			\begin{tabular}{lccc|cccccc}
				\toprule
				Method & DS $\uparrow$ & RC $\uparrow$ & IS $\uparrow$ & Ped $\downarrow$ & Veh $\downarrow$ & Stat $\downarrow$ & Red $\downarrow$ & TO $\downarrow$ & Block $\downarrow$ \\
				\midrule
				1C& $43.2 \pm 6.4$ & $63.2 \pm 7.6$ & $0.71 \pm 0.05$ & $0.01$ & $0.90$ & $0.00$ & $0.05$  & $1.00$ & $0.00$ \\
				3C & $59.6 \pm 0.7$ & $80.6 \pm 7.0$ & $0.73 \pm 0.04$ & $0.01$ & $0.10$ & $0.00$ & $0.02$  & $0.11$ & $0.00$ \\
				3C1A & $68.4 \pm 3.6$ & $80.2 \pm 5.3$ & $0.79 \pm 0.06$ & $\textbf{0.00}$ & $\textbf{0.01}$ & ${0.00}$ & $\textbf{0.00}$ &  $0.07$ & $0.00$ \\
				\midrule
				3C1A1L (M2DA) & $\textbf{72.6} \pm 5.7$ & $\textbf{89.7} \pm 7.8$ & $\textbf{0.80} \pm 0.05$ & $0.01$ & $0.02$ & $0.00$ & $0.01$ &  $\textbf{0.03}$ & $0.00$ \\
				\bottomrule
			\end{tabular}
		}
		\label{ablationsensor}
	\end{table}

	The impact of varying the M2DA architecture is shown in \cref{ablationmodule}. When we remove both the fusion and DA modules, the collision rate with vehicles (Veh) is the highest. After adding the DA module, the model can better capture traffic light information, effectively reducing Veh and Red. It is worth noting that the variance increases when adding the DA module, which can be attributed to the uncertainty caused by the subjective factors of human drivers' visual attention. Upon introducing the LVAFusion, the driving score is significantly enhanced, indicating that the proposed LVAFusion handles multi-modal information effectively and assists the agent in making driving decisions well. As expected, the introduction of both LVAFusion and the DA module results in the best driving performance.
	
	\begin{table}[tb]
		\centering
		\caption{Ablation study for different components of M2DA. $\uparrow$ means the higher, the better, while $\downarrow$ represents the lower, the better. $\checkmark$ represents using the module.}
		\resizebox{\textwidth}{!}{
			\begin{tabular}{clccc|ccccc}
				\toprule
				DA module & Fusion module & DS $\uparrow$ & RC $\uparrow$ & IS $\uparrow$ & Ped $\downarrow$ & Veh $\downarrow$ & Stat $\downarrow$ & Red $\downarrow$  & Block $\downarrow$ \\
				\midrule
				- & - & $51.6 \pm 3.4$ & $88.9 \pm 2.5$ & $0.57 \pm 0.05$ & $\textbf{0.00}$ & $0.09$ & $0.01$ & $0.02$   & $0.07$ \\
				\checkmark & - & $54.1 \pm 11.3$ & $82.8 \pm 10.9$ & $0.64 \pm 0.06$ & $0.03$ & $0.03$ & $0.00$ & $0.01$   & $0.00$ \\
				- & \checkmark & $69.8 \pm 5.6$ & $\textbf{95.1} \pm 4.6$ & $0.72 \pm 0.08$ & ${{0.01}}$ & ${0.06}$ & ${0.00}$ & ${0.01}$ &  $0.01$ \\
				\midrule
				\checkmark & \checkmark & $\textbf{72.6} \pm 5.7$ & $89.7 \pm 7.8$ & $\textbf{0.80} \pm 0.05$  & $0.01$ & $\textbf{0.02}$ & $0.00$ & $0.01$ & $0.00$ \\
				\bottomrule
			\end{tabular}
		}
		\label{ablationmodule}
	\end{table}
	
	\section{Conclusion}

	In this work, we present M2DA, an end-to-end autonomous driving framework focusing on efficient multi-modal environment perception and human-like scene understanding. First, a novel Lidar-Vision-Attention-based Fusion (LVAFusion) module is proposed to fuse multi-modal data better and achieve higher alignment between different modalities. Furthermore, M2DA empowers autonomous vehicles with the human-like scene understanding ability to identify crucial objects by incorporating visual attention information from experienced drivers. After verification, M2DA achieves SOTA performance on two competitive closed-loop autonomous driving benchmarks. 
	
	However, our study has several limitations. First, trajectory prediction, a crucial aspect of autonomous driving, is not meticulously addressed in our model. Instead, we just predict the surrounding vehicles' speed using a sliding window and assume they move at a constant speed to infer their trajectories, which may not accurately capture their multi-modal driving intentions. Additionally, we only investigate single-time-step input data, whereas analyzing time-series input data could provide valuable insights to infer the dynamic states of surrounding objects, potentially leading to improved driving performance.

	\clearpage

	%
	%
	\bibliographystyle{splncs04}
	\bibliography{main}
	
	\newpage
	\appendix
	\section*{Appendix}
	\section{Loss Function Design}
	\label{loss function}
	In order to maintain the stability of the training process, we first independently train the driver attention prediction module with the loss $\mathcal{L}_{da}$. After importing the parameters of the pre-trained DA prediction module, the other modules of M2DA are trained together with a total loss consisting of the waypoint prediction loss $\mathcal{L}_{wp}$ and two auxiliary losses, \ie, perceptual heatmaps loss $\mathcal{L}_{ht}$ and traffic states loss $\mathcal{L}_{tf}$, which can be represented as:
	\begin{equation}
		\mathcal{L} = \lambda_{wp} \mathcal{L}_{wp} + \lambda_{ht} \mathcal{L}_{ht} + \lambda_{tf} \mathcal{L}_{tf}
	\end{equation}
	where $\lambda$ is a hyperparameters to balance the three loss terms. We will introduce these four loss terms in detail in this section.
	
	\subsection{Driver Attention Loss Function}
	For training DA model, the loss $\mathcal{L}_{da}$ is determined based on the predicted saliency map $S$ and its corresponding ground-truth $S^*$. It can be calculated by $\mathcal{L}(S,S^*)$:
	\begin{equation}
		\mathcal{L}(S,S^*) = \lambda_{KLD} \mathcal{L}_{KLD} - \lambda_{CC} \mathcal{L}_{CC} - \lambda_{SIM} \mathcal{L}_{SIM}
	\end{equation}
	where $\lambda$ is a hyperparameters to balance the three loss terms. KLD is Kullback-Leibler divergence, which quantifies the information loss between the probability distribution of the predicted maps and the ground truth, with a smaller value indicating a reduced information loss. $CC \in [-1,1]$ is the Pearson’s correlation coefficient, it computes the linear relationship between random variables in the distributions of predicted saliency map and GT, with a higher value indicating a stronger match between the distributions. $SIM \in [0,1]$ means Similarity between $S$ and $S^*$, larger value means a better approximating. These metrics formulated as:
	
	\begin{equation}
		\mathcal{L}_{KLD}(\boldsymbol{S}, {\boldsymbol{S^*}})=\sum_i^N \boldsymbol{S}(i) \log \left(\varepsilon+\frac{\boldsymbol{S}(i)}{\varepsilon+{\boldsymbol{S^*}}(i)}\right) \\
	\end{equation}
	
	\begin{equation}
		\mathcal{L}_{CC}(\boldsymbol{S}, {\boldsymbol{S^*}})=\frac{\operatorname{cov}(\boldsymbol{S}, {\boldsymbol{S^*}})}{\sigma(\boldsymbol{S}) \sigma({\boldsymbol{S^*}})} \\
	\end{equation}
	
	\begin{equation}
		\mathcal{L}_{SIM}(\boldsymbol{S}, {\boldsymbol{S^*}})=\sum_i^N \min (\boldsymbol{S}(i), {\boldsymbol{S^*}}(i))
	\end{equation}
	where $\varepsilon$ is small number to ensure the stability of numerical calculation. $\operatorname{cov}$ means covariance between $S$  and $S^*$, $i$ represents the index in saliency map.

	\subsection{Waypoint Loss Function}
	Fow waypoint loss $\mathcal{L}_{wp}$, we use $L1$ loss to train our model between predicted waypoints and label named $\mathcal{L}_{wp}$. Our goal is to generate waypoints $w_t$ that closely resemble the waypoint $w_{t}^{gt}$ generated by the expert agent at time-step $t$, the loss function is:
	\begin{equation}
		\mathcal{L}_{wp} = \sum_{t=1}^{T} \left\| w_t - w_t^{gt} \right\|_1
	\end{equation}
	where $T$ denotes the sequence length of the waypoints.

	\subsection{Perception Loss Function}
	The perception information is obtained from the predicted heatmap image $M \in \mathbb{R}^{S \times S \times 7}$, where S is 20. It provides 7 characteristic for potential objects in each grid (existence probability, x, y offset from ego vehicle, width and length of the 2d bbox, speed, yaw). The loss $\mathcal{L}_{ht}$ consists of probability prediction loss $\mathcal{L}_{pro}$ and attributes prediction loss $\mathcal{L}_{attr}$. The perception loss $\mathcal{L}_{ht}$ can be obtained directly by adding $\mathcal{L}_{pro}$   and $\mathcal{L}_{attr}$.
	
	To mitigate the issue of predominantly zero probability predictions caused by sparse positive labels, we follow \cite{shao2023safety}, using a balanced loss function. This loss function calculates the average loss separately for positive and negative labels and then combines them. For the characteristic of the predicted bounding-box (x, y, width, length, speed, yaw), we use $L1$ loss to measure it, which can be described as:
	\begin{equation}
		\mathcal{L}_{\text{attr}} = \frac{1}{\mathbb{S}} \sum_{i} \sum_{j} \sum_{k=1}^{6} \left[1_{\{\tilde{M}_{ij0}=1\}} \left| \tilde{M}_{ijk} - M_{ijk} \right|_1 \right]
	\end{equation}
	where $\tilde{M}_{ij0}$ means the probability of the object at $i^{th}$ row and $j^{th}$ column in ground-truth heatmap $\tilde{M}$. The $k$ from 1 to 6 means x, y offset from ego vehicle, width and length of the 2d bbox, speed and yaw. $\mathbb{S}$ represents the total objects in the heatmap $M \in \mathbb{R}^{S \times S \times 7}$.

	\subsection{Traffic States Loss Function}
	For predicting the traffic information, we further divide $\mathcal{L}_{tf}$ into recognizing the traffic light status $\mathcal{L}_{tl}$ and stop lines $\mathcal{L}_{sl}$, predicting whether it is an intersection $\mathcal{L}_{i}$. The  loss function is:
	\begin{equation}
		\mathcal{L}_{tf} = \lambda_{tl} \mathcal{L}_{tl} + \lambda_{sl} \mathcal{L}_{sl} + \lambda_{i} \mathcal{L}_{i}
	\end{equation}
	where $\lambda$ is a hyperparameters to balance the three loss terms.

	\section{Driver Attention Datasets}
	\label{driver attention dataset}
	\cref{tab:driver attention dataset} presents the various attributes of four publicly accessible driver attention datasets, namely BDD-A \cite{xia2019predicting}, DADA-2000 \cite{fang2019dada}, DReyeVE \cite{alletto2016dr}, and EyeTrack \cite{deng2019drivers}. DReyeVE stands as the first publicly available large-scale driver attention dataset, comprising 555,000 frames extracted from 74 video clips. These video clips were recorded using a roof-mounted camera during naturalistic driving experiments in Italy. Berkeley DeepDrive attention (BDD-A) use 1429 critical scene videos recorded in the U.S. city roads and it is labeled by 45 gaze providers. The traffic videos used in EyeTrack were obtained through a dashcam on urban highways in China and the gaze data were recorded under controlled in-lab conditions with 28 subjects viewing the recorded video clips. Driver attention and driver accident (DADA-2000) use 2000 accidental videos to label the gaze data by 20 gaze providers.
	
	\begin{table}[h!]
		\centering
		\caption{The details of driver attention dataset, we use the four datasets to train our driver attention prediction model.}
		\resizebox{\textwidth}{!}{
			\begin{tabular}{@{}lcccc@{}}
				\toprule
				Character & BDD-A & DADA-2000 & DR(eye)VE & EyeTrack \\
				\midrule
				Frames & 455,787 & 658,476 & 555,000 & 74,825 \\
				Resolution & 1280x720 & 1584x660 & 1920x1080 & 1280x720 \\
				Saliency FPS & 29 & 30 & 25 & 30 \\
				Gaze providers & 45 & 20 & 8 & 28 \\
				Providers per frames & 4 & 5 & 1 & 28 \\
				Scene Sources & City road in the U.S. & Video website & Italy & Urban highway in China \\
				Scene filter & Braking events & Accidents & - & - \\
				Gaze collection & In-lab & In-lab & Natural driving & In-lab \\
				Smoothing filter ($\sigma^2$) & - & 625 & 200 & 3600 \\
				\bottomrule
			\end{tabular}%
		}
		\label{tab:driver attention dataset}
	\end{table}
	
	\section{Controllers}
	We use two PID controllers to obtain throttle, brake, and steer values from the predicted waypoint sequence. Considering the complexity of the traffic system, autonomous vehicles can encounter safety-critical scenarios. In such scenarios, using waypoints alone to control autonomous vehicles may be unsafe. Thus, we adjust the vehicle control signals from the PID controllers by introducing a safety-heuristic method, which can be formulated as: 
	\begin{equation}
		\begin{aligned}
			& \max _{v_d^1} \quad v_d^1 \\
			& \text { s.t. } \quad\left(v_0+v_d^1\right) t \leqslant 2s_1 \\
			& \left(v_0+v_d^1\right) t+\left(v_d^1+v_d^2\right) t \leqslant 2s_2 \\
			& \frac{\left|v_d^1-v_0\right|}{t} \leqslant a_{\text {max }}  \qquad   \frac{\left|v_d^2-v_d^1\right|}{t} \leqslant a_{\max } \\
			&
		\end{aligned}
	\end{equation}
	where $v_0$ denotes the current velocity of the ego vehicle, and $v_d^1$ and $v_d^2$ are the velocity of the next 0.5s and 1.0s, respectively. The $s_1$ and $s_2$ are the maximum safe distance of the next 0.5s and 1.0s, respectively. The goal of the safety-heuristic method is to maximize the agent's traffic efficiency while ensuring safety.
	
	\section{Implementation Details}
	\label{Implementation}
	We use 2 sensor modalities including three RGB cameras ($60^o$ left, forward and $60^o$ right) and one Lidar sensor.  All cameras have the resolution of 800 × 600 and a horizontal field of view (FOV) of $100^o$. Because of the distortion caused by the rendering of the cameras in CARLA simulator, the front view of the image is cropped to 3 × 224 × 224. For the left and right views image, we crop them to 3 × 128 × 128 . To improve the model's comprehension of complex environments, we aim to learn the attention patterns of human drivers. We adopt DA prediction model to get the driver's gaze, and consider it as a mask to modify the weight of the raw image, crop it to 3 × 224 × 224.
	
	For Lidar point clouds, we follow previous works \cite{rhinehart2019precog, shao2023safety, chitta2022transfuser, gaidon2016virtual} to convert the Lidar point cloud data into a 2D BEV grid map by calculating the number of Lidar points inside each grid. We consider the area of the 2D BEV grid map is 32×32m, with 28m in front of the vehicle, 4m behind the vehicle, and 16m to each of the sides. We partition the grid into 0.125m × 0.125m cells, yielding a resolution of 256 × 256 pixels. We used random scaling from 0.9 to 1.1 and color jittering for data augmentation.
	
	For training the DA model, we following \cite{gan2022multisource}, we initialized the learning rate at 0.02 and decayed it exponentially by a factor of 0.8 after each epoch. We used stochastic gradient descent with a momentum of 0.9 and applied weight decay of $10^-4$ for optimization. Then we use the trained weight to predict the driver attention.
	
	As for model architecture, we adopt a pretrained Resnet50 model as the backbone for encoding the information from multi-view RGB images and salience image, employ the pretrained Resnet18 model as the backbone to extract the Lidar features. We utilize the output of stage 4 in a standard Resnet as the tokens to the downstream fusion module.  As shown in LVAFusion, we design 2 cross-attention mechanisms to interact with point clouds and images respectively, and adopt a self-attention to enhance the extracted features. The dimension of the features is 256. The layer $K$ of transformer is 6. 
	
	We import partial trained weights of  \cite{shao2023safety} as the initial weights for M2DA (only for the compatible modules between two models and freeze them) and train our model with 8 Tesla V100 GPUs for 35 epoch, with an initial learning rate 5$e^{-4}$ for transformer and 2$e^{-4}$ for Resnet backbone. The batchsize is 16 per GPU card. We use the AdamW optimizer \cite{loshchilov2017decoupled} and cosine learning rate scheduler \cite{loshchilov2016sgdr}. The details of $\lambda$ and other hyperparameters for training is shown in \cref{tab:learning_process}

	\begin{table}[htb]
		\centering
		\caption{Details of the $\lambda$ and other other hyperparameters in M2DA.}
		\label{tab:learning_process}
			\begin{tabular}{@{}lll@{}}
				\toprule
				Notation & Description & Value \\ \midrule
				\( \lambda_{wp} \) & Weight for the waypoints loss & 0.8 \\
				\( \lambda_{ht} \) & Weight for the heatmap loss & 1.0 \\
				\( \lambda_{tf} \) & Weight for the traffic states loss & 0.8 \\
				\( \lambda_{KLD} \) & Weight for the Kullback-Leibler divergence loss & 0.9 \\
				\( \lambda_{CC} \) & Weight for the Pearson’s correlation coefficient loss & 0.1 \\
				\( \lambda_{SIM} \) & Weight for the Similarity loss & 0.1 \\
				\( \lambda_{tl} \) & Weight for the traffic light loss & 0.5 \\
				\( \lambda_{sl} \) & Weight for the stop sign loss & 0.1 \\
				\( \lambda_i \) & Weight for the intersection loss & 0.1 \\
				\( \text{Threshold} \) & Threshold for filtering objects in heatmap & 0.9 \\
				\( a_{max} \) & Maximum acceleration for agent & 1.0$m/s^2$ \\
				\( v_{max} \) & Maximum velocity for agent & 5.0$m/s$ \\
				\(M \) & Size of the heatmap & $20 \times 20$ \\
				\( \text{Collision buff} \) & Mnimum collision safety distance & [3.7, 2] \\
				\bottomrule
			\end{tabular}%
	\end{table}

	\section{Benchmark details}
	\label{benchmark details}
	\subsection{Carla Town05 Long Benchmark}
	We use Town05 for evaluation and other towns for training. 
	In the Town05 Long benchmark, it has 10 long routes of 1000-2000m, details can be seen in \cref{tab:town5route_info}, each comprising 10 intersections. and Town05 has a wide variety of road types, including single-lane roads, bridges, highways. 
	
	\begin{table}[htb]
		\caption{Detailed Route Information about Town05 Long benchmark.}
		\centering
		{%
			\begin{tabular}{ccccccccccc}
				\toprule
				Index & 16 & 17  & 18 & 19 & 20 & 21 & 22  & 23 & 24 & 25 \\
				\midrule
				Length (m) & 1071 & 862 & 1018 & 1650 & 1247 & 531 & 991 & 1271 & 2101 & 1554 \\
				\bottomrule
			\end{tabular}%
		}
		\label{tab:town5route_info}
	\end{table}
	
	\subsection{Carla Longest6 Benchmark}
	Longset6 benchmark has 36 routes with an average route length of 1.5km, which is similar to the average route length of the official leaderboard (1.7km), and it has a high density of dynamic agents. Moreover, each route has a unique environmental condition. The details of Longest6 is shown in \cref{tab:route_info}. The core challenge of the benchmark is how to handle dynamic agents and adversarial events.
	\begin{table}[htb]
		\caption{Detailed Route Information about Longest6 benchmark.}
		\centering
		\resizebox{\textwidth}{!}{%
			\begin{tabular}{ccccc|ccccc|ccccc}
				\toprule
				Route & Town & Weather  & Daytime & Length & Route & Town & Weather  & Daytime & Length & Route & Town & Weather  & Daytime & Length \\
				\midrule
				0 & 1 & MidRain & Dawn & 1130 & 12 & 3 & SoftRain & Twilight & 2303 & 24 & 5 & MidRain & Sunset & 2101 \\
				1 & 1 & Cloudy & Dawn & 1014 & 13 & 3 & MidRain & Twilight & 1748 & 25 & 5 & SoftRain & Noon & 2554 \\
				2 & 1 & Cloudy & Morning & 893 & 14 & 3 & WetCloudy & Night & 1436 & 26 & 5 & SoftRain & Morning & 1271 \\
				3 & 1 & HardRain & Noon & 731 & 15 & 3 & MidRain & Noon & 1870 & 27 & 5 & WetCloudy & Morning & 1078 \\
				4 & 1 & HardRain & Twilight & 636 & 16 & 3 & HardRain & Night & 1856 & 28 & 5 & MidRain & Morning & 1071 \\
				5 & 1 & HardRain & Morning & 985 & 17 & 3 & Wet & Dawn & 1569 & 29 & 5 & WetCloudy & Dawn & 1651 \\
				6 & 2 & Wet & Noon & 1010 & 18 & 4 & WetCloudy & Twilight & 2069 & 30 & 6 & SoftRain & Dawn & 2525 \\
				7 & 2 & Cloudy & Night & 974 & 19 & 4 & SoftRain & Dawn & 2058 & 31 & 6 & Wet & Dawn & 1859 \\
				8 & 2 & Cloudy & Twilight & 820 & 20 & 4 & SoftRain & Night & 1862 & 32 & 6 & Wet & Twilight & 2842 \\
				9 & 2 & WetCloudy & Noon & 920 & 21 & 4 & HardRain & Night & 1863 & 33 & 6 & Wet & Night & 2270 \\
				10 & 2 & HardRain & Sunset & 872 & 22 & 4 & Cloudy & Sunset & 2319 & 34 & 6 & Cloudy & Noon & 1442 \\
				11 & 2 & MidRain & Night & 872 & 23 & 4 & Wet & Sunset & 2440 & 35 & 6 & WetCloudy & Sunset & 1760 \\
				\bottomrule
			\end{tabular}%
		}
		\label{tab:route_info}
	\end{table}

	\section{Results and Visualizations}
	\begin{table}[t]
		\centering
		\caption{Comparison of M2DA with several state-of-the-art methods in the Longest6 benchmark. $\uparrow$ denotes the higher, the better, while $\downarrow$ represents the lower, the better. The details of Infraction Score are displayed. Ped denotes collision with pedestrians. Veh means collision with vehicles. Stat represents collision with static objects. Red is the red light violation. TO denotes time out. Block means the agent is blocked. The evaluation result of the expert agent comes from \cite{chitta2022transfuser}.}
		\resizebox{\textwidth}{!}{
			\begin{tabular}{llccc|cccccc}
				\toprule
				Method & Dataset & DS $\uparrow$ & RC $\uparrow$ & IS $\uparrow$ & Ped $\downarrow$ & Veh $\downarrow$ & Stat $\downarrow$ & Red $\downarrow$ & TO $\downarrow$ & Block $\downarrow$ \\
				\midrule
				WOR \cite{chen2021learning} & 1M & $20.5 \pm 3.1$ & $48.5 \pm 3.9$ & $0.56 \pm 0.03$ & $0.18$ & $1.05$ & $0.37$ & $1.28$  & $0.08$ & $0.20$ \\
				LAV \cite{chen2022learning} & 189K & $32.7 \pm 1.5$ & $70.4 \pm 3.1$ & $0.51 \pm 0.02$ & $0.16$ & $0.83$ & $0.15$ & $0.96$  & $0.12$ & $0.45$ \\
				Interfuser \cite{shao2023safety} & 3M & $47.0 \pm 6.0$ & $74.0 \pm 1.0$ & $0.63 \pm 0.07$ & $0.06$ & $1.14$ & $0.11$ & $0.24$ &  $0.52$ & $0.06$ \\
				TransFuser \cite{chitta2022transfuser} & 228K & $47.3 \pm 5.7$ & $\textbf{93.1} \pm 1.0$ & $0.50 \pm 0.06$ & $0.03$ & $2.45$ & $0.07$ & $0.16$ & $\textbf{0.06}$ & $0.10$ \\
				TCP \cite{wu2022trajectory} & 189K & $48.0 \pm 3.0$ & $72.0 \pm 3.0$ & $0.65 \pm 0.04$ & $0.04$ & $1.08$ & $0.23$ & $0.14$ & $0.18$ & $0.35$ \\
				P- PlanT\cite{renz2022plant} & 228K & $58.0 \pm 5.0$ & $88.0 \pm 1.0$ & $0.65 \pm 0.06$ & $0.07$ & $0.97$ & $0.11$ & $0.09$ &  $0.13$ & $0.13$ \\
				\textbf{M2DA} (ours) & 200K & $\textbf{64.5} \pm 3.1$ & $85.2 \pm 3.6$ & $\textbf{0.76} \pm 0.01$ & $\textbf{0.01}$ & $\textbf{0.12}$ & $\textbf{0.05}$ & $\textbf{0.07}$ &  $0.23$ & $\textbf{0.04}$ \\
				\midrule
				Expert & None & $76.9 \pm 2.2$ & $88.7 \pm 0.6$ & $0.86 \pm 0.0$ & $0.02$ & $0.28$ & $0.01$ & $0.03$ &  $0.08$ & $0.13$ \\
				\bottomrule
			\end{tabular}
		}
		\label{longest6resullt}
	\end{table}
	
	Transfuser proposes the Longest6 benchmark and exhibits the highest road completion rate (\cref{longest6resullt}). However, its aggressive driving style also records the highest vehicle collision rate, leading to a reduction in the driving score. By contrast, M2DA performs much better in DS. Meanwhile, M2DA also obtains the highest IS. In addition to DS, RC, and IS, we also present the infraction details for comprehensive analysis. Compared with its teacher \ie, the expert agent, M2DA achieves a lower collision rate in Ped and Veh, implying that M2DA performs well when confronting complex traffic scenarios with randomly generated pedestrians or vehicles. By avoiding collisions with pedestrians and vehicles, M2DA demonstrates safer driving capability, which is crucial for the actual widespread application of autonomous driving. Moreover, since we design a prediction head that can provide traffic light status as an additional decision reference, M2DA exhibits the lowest probability of running a red light (Red).

	\begin{figure}[tb]
		\label{pic2}
		\centering
		\includegraphics[scale=0.32]{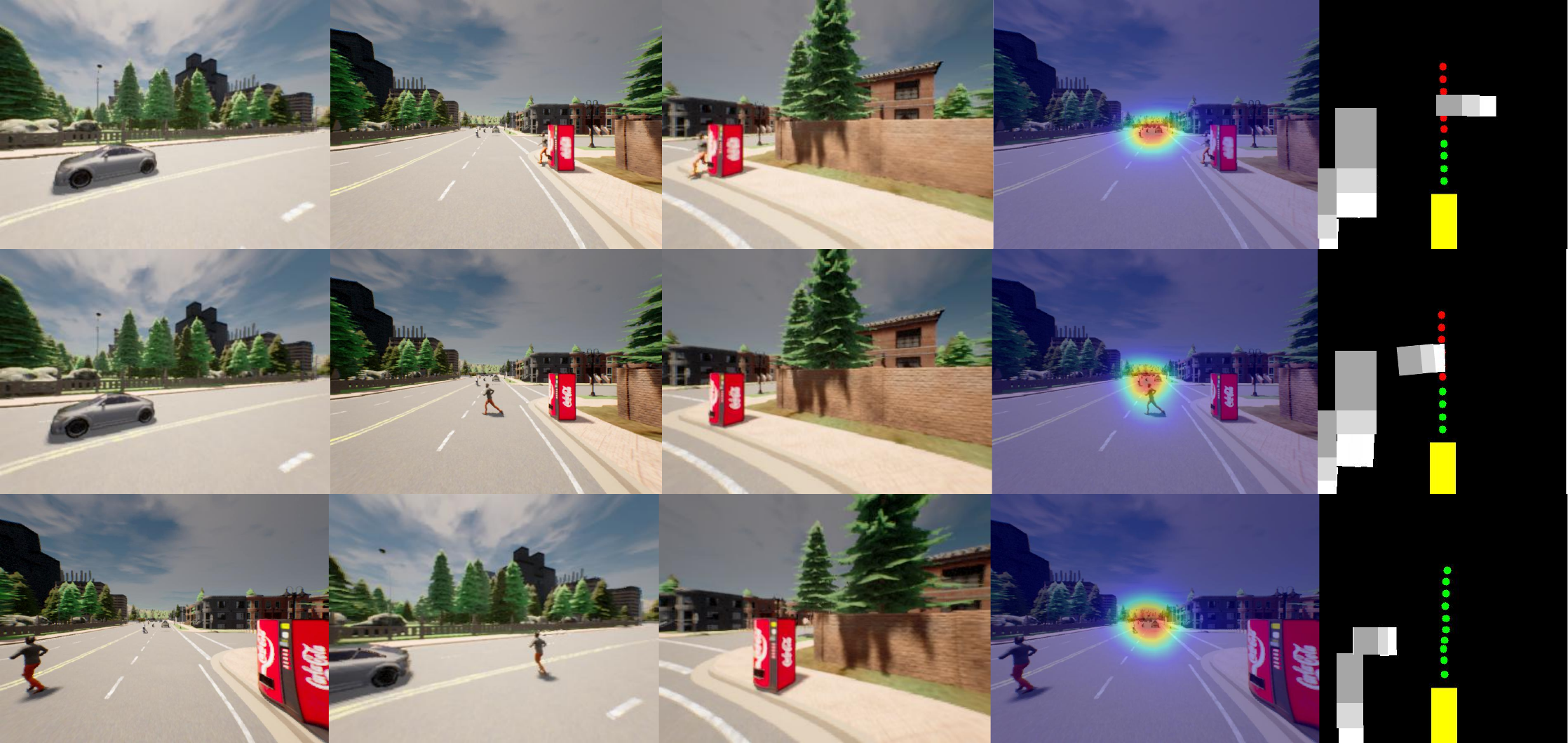}
		\caption{Detailed visualization of the pedestrian crossing case. 
		}
		\label{fig:pic2}
	\end{figure}
	
	\begin{figure}[t!]
		\label{wrong}
		\centering
		\includegraphics[scale=0.32]{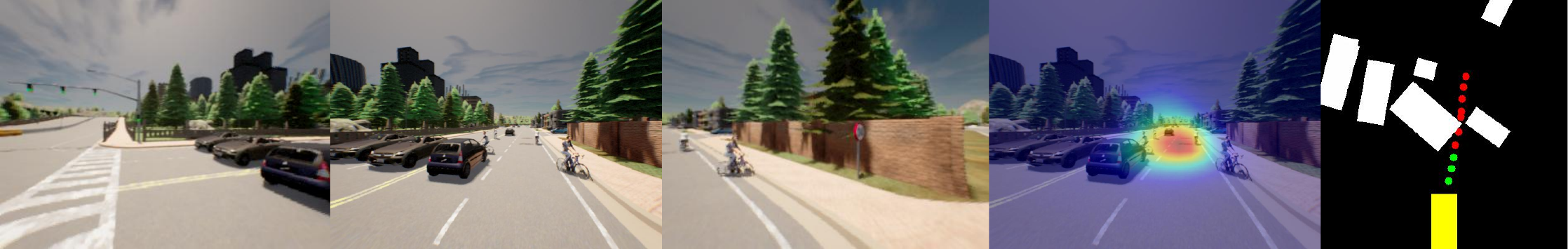}
		\caption{Visualization of a failure case with three RGB images, the predicted driver attention, and a heatmap image representing perceptual information. Yellow and white boxes denote the ego vehicle and perceived surrounding objects, respectively. Green dots and red dots represent safe future trajectories and unsafe areas where collisions are likely to occur, respectively.
		}
		\label{fig:wrong}
	\end{figure}
	
	We visualize the more details of the pedestrian crossing case in the evaluation results of M2DA (\cref{fig:pic2}). The first row displays a normal traffic scenario without apparent risks, where M2DA located its visual attention at the road vanishing point in the center of the image. In the second row, a running pedestrian was about to cross the road. In such a sudden situation, like an experienced human driver, M2DA quickly and accurately captured the dangerous object \ie, the pedestrian, in the current traffic scenario and made corresponding driving decisions to avoid potential collisions. After pedestrians cross the road, M2DA realigns its focus on the road ahead, thereby enhancing the interpretability of the decision-making process.
	
	\label{Visualizations}
	In \cref{fig:wrong}, we also provided a failure case of M2DA. Due to the prediction errors of the surrounding vehicle's yaw, its future trajectory was incorrectly estimated. Consequently, the ego vehicle misperceived an ongoing danger ahead and thus remained stationary, unable to proceed forward as anticipated.  
	
\end{document}